\documentclass{article}

\usepackage{PRIMEarxiv}

\usepackage[utf8]{inputenc}    % For UTF-8 encoding
\usepackage[T1]{fontenc}       % For font encoding
\usepackage{hyperref}          % For hyperlinks
\usepackage{url}               % For URL formatting
\usepackage{booktabs}          % For professional-quality tables
\usepackage{amsfonts}          % For additional math fonts (if needed)
\usepackage{nicefrac}          % For compact fractions (optional)
\usepackage{microtype}         % For microtypography improvements
\usepackage{graphicx}          % For including images
\usepackage{tabularx}          % For flexible width tables
\usepackage{fancyhdr}          % For custom headers
\usepackage{lipsum}            % For placeholder text (optional, remove if unnecessary)
\usepackage{amsmath}

\graphicspath{{media/}}     % organize your images and other figures under media/ folder

\title{Token Pruning using a Lightweight Background Aware Vision Transformer} 

\author{
  {\bf Sudhakar Sah, Ravish Kumar, Honnesh Rohmetra, Ehsan Saboori} \\ \\
  Deeplite\\
  Toronto, Canada \\
  \texttt{sudhakar@deeplite.ai}
}
\begin{document}

\maketitle

\begin{abstract}
High runtime memory and high latency puts significant constraint on Vision Transformer training and inference, especially on edge devices. Token pruning reduces the number of input tokens to the ViT based on importance criteria of each token. We present a Background Aware Vision Transformer (BAViT) model, a pre-processing block to object detection models like DETR/YOLOS aimed to reduce runtime memory and increase throughput by using a novel approach to identify background tokens in the image. The background tokens can be pruned completely or partially before feeding to a ViT based object detector.  We use the semantic information provided by segmentation map and/or bounding box annotation to train a few layers of ViT to classify tokens to either foreground or background. Using 2 layers and 10 layers of BAViT, background and foreground tokens can be separated with $75$\% and $88$\% accuracy on VOC dataset and $71$\% and $80$\% accuracy on COCO dataset respectively. We show a 2 layer BAViT-small model as pre-processor to YOLOS can increase the throughput by $30$\% - $40$\% with a mAP drop of $3$\% without any sparse fine-tuning and $2$\% with sparse fine-tuning. Our approach is specifically targeted for Edge AI use cases. Code and data are available at~[Link].  

  % \keywords{ViT \and Sparsity \and Token Pruning \and Sparse-ViT \and Background subtraction}
\end{abstract}

\section{Introduction}
\label{sec:intro} Transformers \cite{transformers} have already demonstrated their ability to outperform traditional methods in Natural Language Processing (NLP) with models like BERT \cite{bert} and RoBERTa \cite{roberta}. They are now commonly used in modern vision-related tasks such as classification \cite{swin}, object detection \cite{detr} \cite{wbdetr} \cite{yolos}, segmentation \cite{semantic_segmentation}, and pose estimation \cite{ViT_pose} as Vision Transformers(ViT). 
Despite the advantages of ViTs over traditional CNN-based approaches, their high computational requirements pose significant challenge in deployment of these models on edge devices with limited memory and computational power. The ViT accepts small image patches (typically $16 \times 16$ size) called \textit{tokens} as input. As image resolution increases, more input tokens are generated, which increases the performance but reduces model throughput and latency. \\ ViT is also used for object detection by models like DETR \cite{detr} which uses learnable queries and encoder features to produce box predictions using decoder. Different variations of DETR-like models like \cite{rtdetr}\cite{wbdetr}, \cite{yolos} are proposed to create state of the art object detection models. 

\begin{figure}
    \centering
    \includegraphics[width=0.8\linewidth]{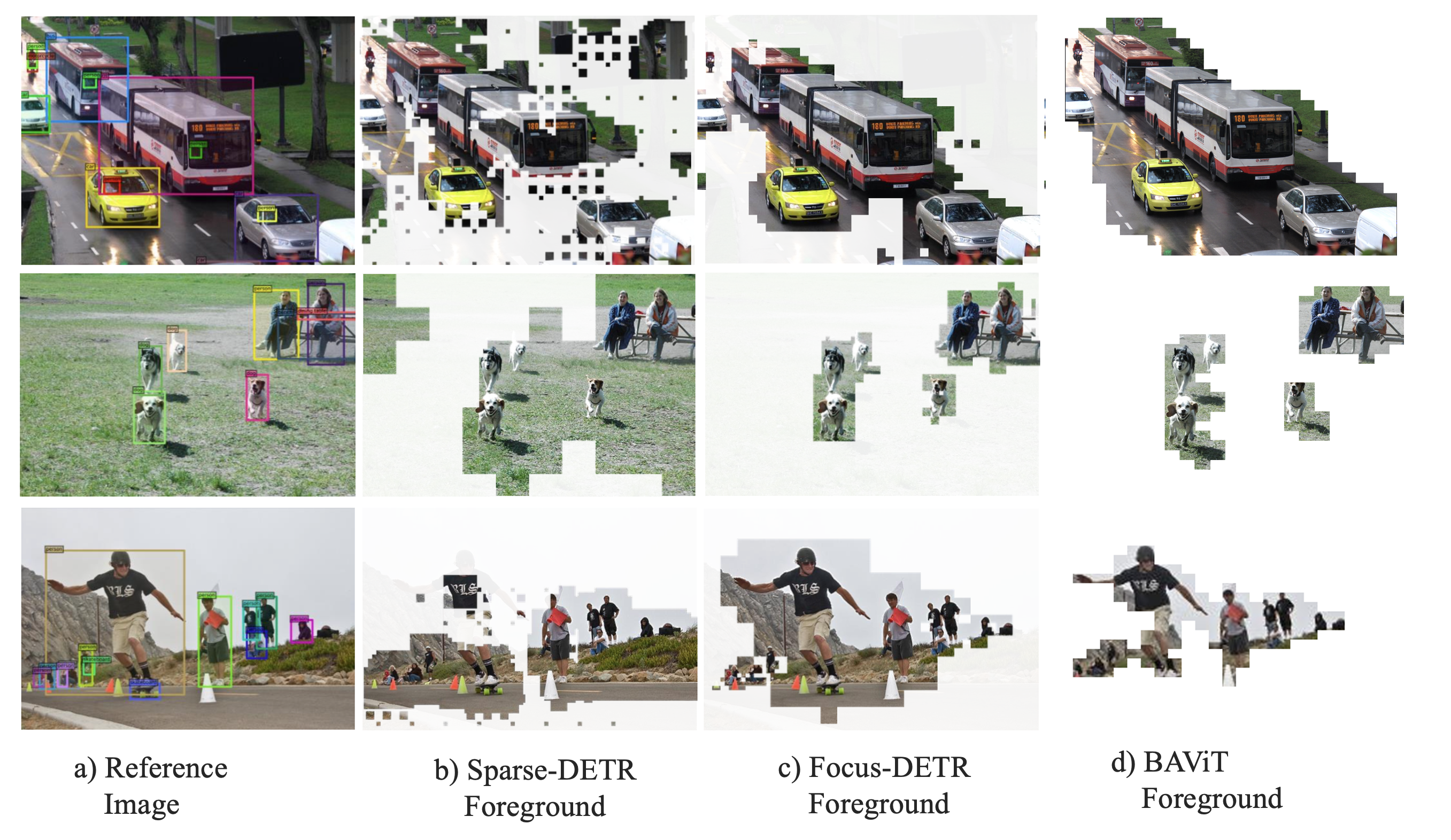}
    \caption{Comparison of background token identification results between Sparse DETR, Focus DETR and BAViT models}
    \label{fig:bavit_vs_others}
\end{figure}

Zheng et.al \cite{focus_detr} showed that the complexity of Deformable DETR \cite{deformable_detr} is $8.8\times$ compared to the decoder which suggests that focusing on efficiency of the encoder is very important. All the tokens do not have same importance and by reducing the number of tokens results in latency and throughput improvement. The technique to reduce the number of tokens by assessing the importance or relevance of each token is called \textit{token pruning}. In this work, we aim to reduce the number of input tokens by introducing a novel token importance criteria for pruning with a minimal impact on performance. Our approach uses segmentation masks provided in the COCO (80 object categories) \cite{lin2014microsoft} and PASCAL VOC (20 object categories) \cite{pascal-voc-2012} datasets to annotate each individual patch as foreground (FG) or background (BG). This annotation serves as a guide for ViT models in object detection tasks to determine the importance of each token. Sparse DETR \cite{sparse_detr} and Focus DETR \cite{focus_detr} are two most impressive and state of the art techniques for token pruning. As show in Figure \ref{fig:bavit_vs_others}, sparse DETR uses the token importance score by computing cross-attention map in the decoder which reduces the number of tokens by 70\%. Focus DETR \cite{focus_detr} on the other hand detects background tokens and prunes them to increase the throughput. We propose a method that uses background token detection similar to focus DETR but we our target usecase is Edge devices so we avoid using heavy CNN backbones, as proposed in Focus DETR, to detect background tokens which could be computationally very expensive. 

We summarize our contributions as follows:
\begin{itemize}
    \item We introduce a novel Background Aware Vision Transformer ({\em BAViT}) capable of separating FG and BG tokens
    \item We introduce a modified Accumulative Cross Entropy Loss function for BG/FG classification. 
    \item We demonstrate that integration of {\em BAViT} as pre-processing block of DETR/YOLOS like object detection model provides a good latency/accuracy trade-off and increased throughput of the model. 
\end{itemize}

\section{Related Work}
\label{sec:related_work}
\subsection{Vision Transformers}
Transformers \cite{transformers} have emerged as a dominant architecture in NLP \cite{bert} \cite{roberta} as well as vision-related tasks \cite{rtdetr} \cite{yolos} and these models \cite{vit} have achieved state-of-the-art performance for vision tasks including object detection such as DETR  \cite{detr}, RT-SETR\cite{rtdetr} and YOLOS\cite{yolos}. DETR \cite{detr} employs a combination of CNN-based backbones followed by transformers to address object detection tasks. Swin-transformers \cite{swin} introduced new ViTs that can serve as general-purpose backbones for computer vision tasks. WBDetr\cite{wbdetr} replaced the CNN-based backbones in DETR \cite{detr} with a transformer-based backbone for object detection. Similarly, innovations continue to enhance ViT capabilities, such as \cite{all_tokens_matter}, which introduces a K-dimensional score map to provide localized information about image patches. Recent work by Fang et al. \cite{yolos} proposes end to end object detection as sequence-to-sequence task. Our BAViT proposes an additional information about of these image tokens as BG and FG, which can be integrated as the pre-processing stage to filter out unnecessary patches.

\subsection{Runtime Memory Improvement}
ViTs \cite{vit} require substantial runtime memory, which limits their use on smaller devices. Many research efforts, including \cite{sparse_attention} \cite{sparse_sinkhorn_attention} \cite{big_bird}, propose methods to optimize the performance of vision transformers. Reformer \cite{reformer} introduces architectural changes to the residual layers, replacing them with reversible residual layers to make the model more efficient. Sparse attention\cite{sparse_attention} proposes an alternative attention formulation through sparse factorization of the attention matrix, which is one of the most computationally expensive components in ViTs. Sparse Detr\cite{sparse_detr} enhances the efficiency of DETR \cite{detr}-like models by substituting dense attention with deformable attention. Other works, such as \cite{dropout}, replace standard dropout layers with structured dropout layers to improve the efficiency and robustness of transformers. Few methods focus on pruning heads by ranking them based on their estimated importance \cite{head_pruning}. Additionally, quantization approaches \cite{8_bit_quantize} \cite{hessian_quantize} \cite{noise_quantize} have been explored to further improve the efficiency of ViTs.

\subsection{Token Pruning}
The number of tokens contribute to quadratic complexity in ViTs during inference. However, all the tokens generated from the input image are not equally important; many primarily contain background information. Several research efforts, including \cite{dynamic_sparse_vit} \cite{revisiting_token_pruning} \cite{adaptive_sparse_vit} \cite{sparse_detr} \cite{focus_detr}, propose efficient approaches to remove unnecessary tokens, thereby improving the inference time. \cite{focus_detr} introduces a technique that efficiently scores the importance of tokens, discards background queries, and enhances the semantic interaction of fine-grained object queries based on these scores. \cite{adaptive_sparse_vit} proposes an adaptive method to hierarchically discard useless tokens and adjust computational costs for different input instances. \cite{revisiting_token_pruning} suggests reusing pruned tokens at later stages of the model. Our work is very close to Focus DETR \cite{focus_detr} as both approaches focus on classifying tokens into FG and BG. However, Focus DETR uses a heavy backbone from DETR (like ResNet50, ResNet101 \cite{resnet}) which is not suitable for edge devices. Also, Focus DETR proposes many modifications to the existing DETR model which requires model retraining or fine-tuning for a long time. Therefore, although the technique produces SOTA results, it is not feasible approach for edge devices. Our work proposes a simpler strategy for background token identification using a learnable small ViT model using 2 layers. Also, our approach produces foreground images which visibly looks very similar to Focus DETR produced foreground images but our approach uses a very small model, compared to Focus DETR, to achieve this. BAViT can be used as a separate module and integrated with other models at the pre-processing data stage, enabling faster performance and making the models suitable for smaller devices. Our target use case is small ViTs for edge devices, therefore it is difficult to compare our method with Focus DETR mAP/latency numbers which uses very large model and performs latency experiments on larger GPUs. 

\section{Methodology}
\label{sec:methodology}
\subsection {Auxiliary Annotations} 
\begin{figure}
    \centering
    \includegraphics[width=1.0\linewidth]{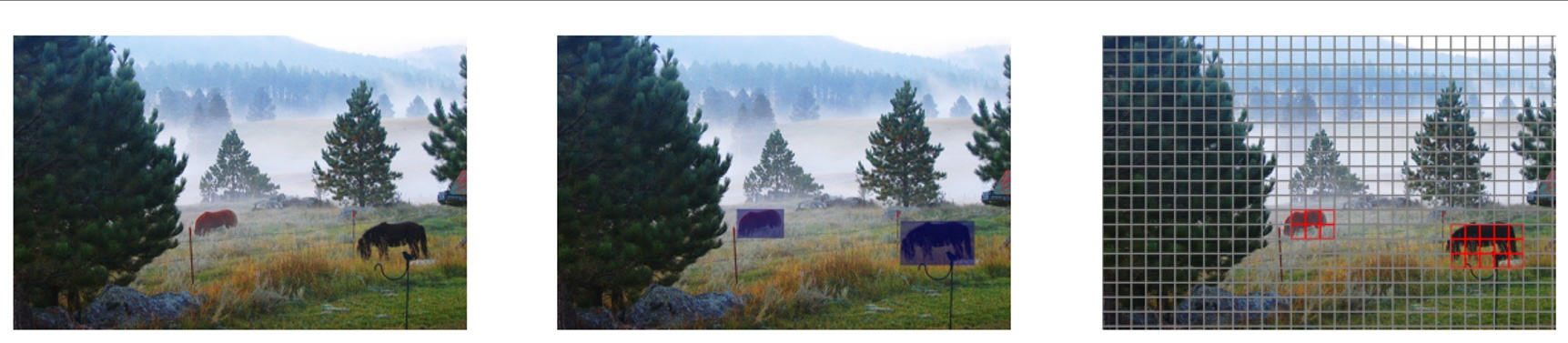}
    \caption{Three VOC images (left to right) a) original image b) foreground object area in transparency c)$16\times16$ grid with red grids bring foreground and gray being background}
    \label{fig:fig_annotation}
\end{figure}

Transformers accept image patches (called tokens) of size ($k \times k$), created by dividing the input image into a sequence of square patches, as shown in Figure \ref{fig:fig_annotation}. ViTs use these patches to classify objects in the image through the attention mechanism. Popular datasets like Microsoft COCO \cite{lin2014microsoft} and Pascal VOC \cite{pascal-voc-2012}, used for object detection and segmentation tasks, contain annotations such as bounding boxes and instance segmentation maps. We create a M-dimensional patch annotation vector for every input image, where \textit{M} represents the total number of tokens formed by dividing the input image into $k \times k$ smaller non-overlapping patches as shown in Figure \ref{fig:fig_annotation}. We compare the Jaccard similarity coefficient \cite{rezatofighi2019generalizedintersectionunionmetric}  of each token with all the bounding boxes or segmentation map and it is labeled as one (Foreground - FG) if the overlap of a token with any of the bounding box is more than 0.5, otherwise it is labeled as zero (Background - BG) as shown in Equation \ref{eq:label_assign} and Equation \ref{eq:label_assign_jaccord}. Figure \ref{fig:fig_annotation} shows a sample Pascal VOC image (left), bounding boxes (center), and image patches with BG patches in gray and FG patches in red color. When using segmentation maps to create the annotation vector, any image patch with more than {10}\% overlapping pixel with any class of segmentation map is considered foreground; otherwise, it is considered background. We trained BAViT model both using bounding  box annotations and segmentation maps but most of the results presented in this paper are from annotated data using segmentation map. 
\begin{equation}
\label{eq:label_assign}
\textbf{L}_{\textbf{i}} = 
\begin{cases} 
1 & \left\{ \text{if } J(P_i, B_j) \geq \tau \right. \\
0 & \left\{ \text{if } J(P_i, B_j) < \tau \right.
\end{cases}  
\end{equation}

\begin{equation}
\label{eq:label_assign_jaccord}
\textbf{J}(\textbf{P}_i, \textbf{B}_j) = \frac{|P_i \cap B_j|}{|P_i \cup B_j|}
\end{equation}

where \(P_{i}\) is patch and \(B_{i}\) is bounding box, \(L_{i}\) is assigned label for \(i^{\text{th}}\) token, \(\textbf{J}(P_i, B_j)\) is Jaccard coefficient, \(\tau\) is threshold for selecting the token as foreground or background. 

\subsection {BAViT Architecture}

BAViT architecture is created by introducing few fundamental changes in the traditional ViT architecture as illustrated in Figure \ref{fig:bavit_methodology} (left). We remove the CLS token and introduce a linear layer with two output classes for each token. Traditional ViT uses CLS token to encapsulate knowledge from all tokens and it provides the score for each class. On the contrary, BAViT calculates classification score for FG and BG classes for each token. Therefore, we do not need a CLS token. Accumulative Cross Entropy Loss ($L_{acc}$) is calculated as defined in equation \ref{eq:acel}, and weights are updated via back propagation. This loss function can also be used with other loss functions targeting different vision tasks, such as object detection loss to help the model focus on important tokens. Since BAViT is supposed to be used as pre-processing step for token pruning, we decided to keep it light weight and used the model with only 2 layers (BAViT-small) to study the impact on YOLOS \cite{yolos}. However, we have provided BG/FG classification results with 10 layers as well (BAViT-large) as BAViT-small in result section to show the scalability and flexibility of this approach. 

\subsection{Accumulative Cross Entropy Loss}

In contrast to the traditional ViT classifier training, which involves introducing an additional classification token (CLS) and calculating loss only for that token, we propose a new loss function that calculates the Cross Entropy Loss \cite{mao2023crossentropylossfunctionstheoretical} for each token individually and then aggregates these losses. This aggregated loss is termed as Accumulative Cross Entropy Loss ($L_{acc}$), as defined in \ref{eq:acel}. 

% Additionally, we define the average loss across all tokens as Mean Cross Entropy Loss ($L_{mean}$), as defined in \ref{eq:mcel} which can be used for large number of tokens.

% Accumulative Cross-Entropy Loss
% \begin{equation}
% \label{eq:acel}
% L_{acc} = -\sum_{i=1}^{N} \sum_{j=1}^{M} \sum_{c=1}^{C} y_{i,j,c} \log(\hat{y}_{i,j,c})
% \end{equation}

\begin{equation}
\label{eq:acel}
\textbf{L}_{\textbf{acc}} = -\frac{1}{N \times M} \sum_{i=1}^{N} \sum_{j=1}^{M} \sum_{c=1}^{C} y_{i,j,c} \log(\hat{y}_{i,j,c})
\end{equation}

where \( N \) is the the number of image samples, \( M \) is the number of tokens per sample, \( C \) is the number of classes (background and foreground).  \( y_{i,j,c} \) is the variable indicating whether the \( j \)-th token in the \( i \)-th sample belongs to class \( c \). It's value is one if the token belongs to class \( c \), otherwise it is zero. \( \hat{y}_{i,j,c} \) is the predicted probability of the \( j \)-th token in the \( i \)-th sample being in class \( c \). 

\subsection{Model Training}

\begin{figure}
    \centering
    \includegraphics[width=1.0\linewidth]{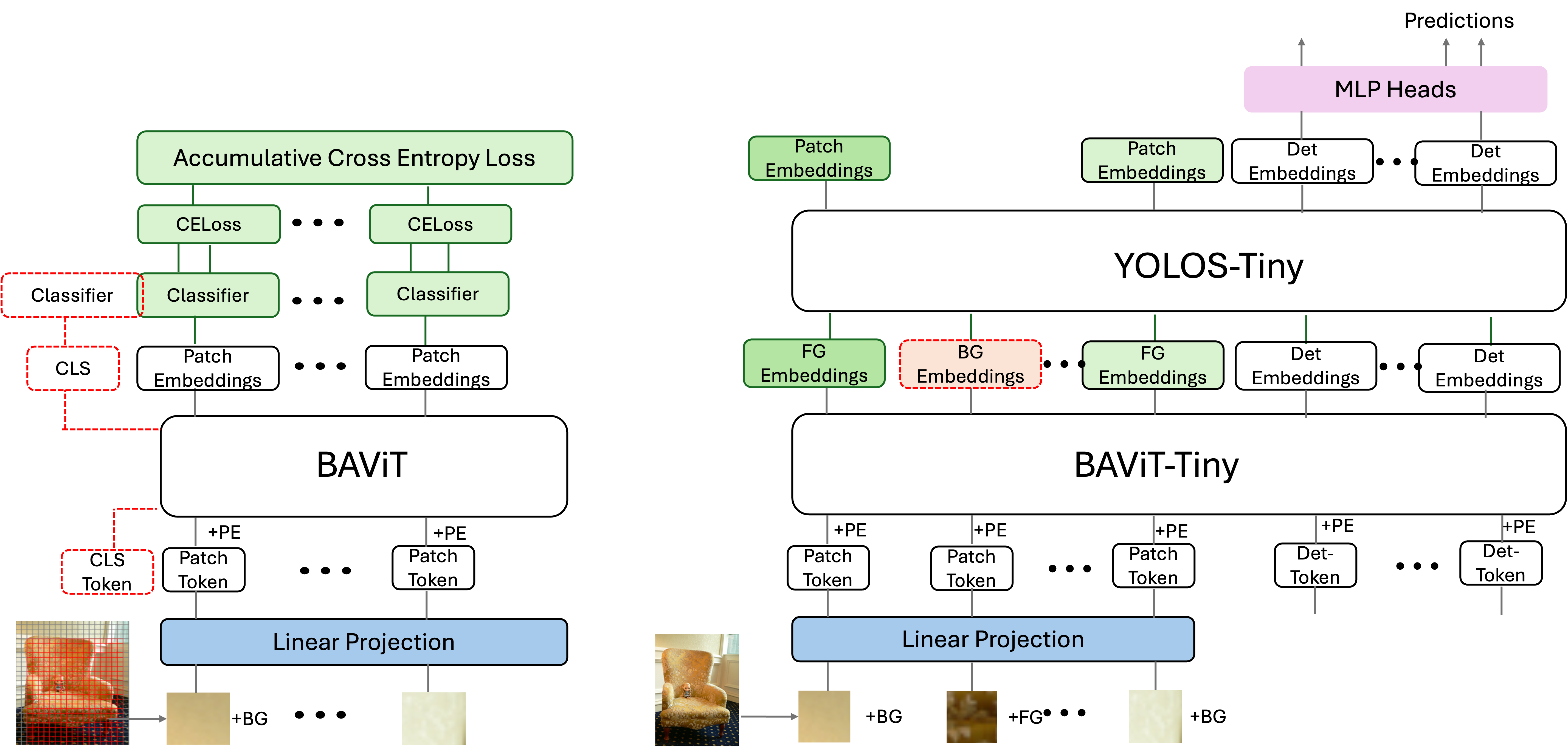}
    \caption{Background aware ViT architecture. (Left) 2 layers for BG and FG patch classification. (Right) BAViT attached as a pre-processing step to YOLOS (DETR type object detector) object detector}
    \label{fig:bavit_methodology}
\end{figure}
We use both Pascal VOC \cite{pascal-voc-2012} and COCO 2017 \cite{lin2014microsoft} to train BAViT and reported mAP (mean Average Precision) result on the validation dataset for both. Each training batch, denoted as ($B$, $M$, $S$), consists of $M$ tokens of size $16 \times 16$, each with an embedding size of $S = 192$, and labels for each token indicating either \textit{BG} or \textit{FG}. We employed the Adam \cite{kingma2017adammethodstochasticoptimization} optimizer with a step learning rate scheduler and trained the model for 100 epochs until convergence. The initial weights for the ViT \cite{googlresearch_vit} model were loaded from ImageNet-1k \cite{deng2009imagenet} dataset pre-trained model.

\subsection {BAViT Integration with ViT based Detection}
\label{sub:bavit_integration}
The BAViT-small is added as a pre-processing block of the ViT based object detector as shown in Figure \ref{fig:bavit_methodology}. We have used YOLOS \cite{yolos} as the object detection model , an architecture similar to DETR\cite{detr} with an exception that YOLOS provides an option to use the detector without a CNN backbone. Our method works directly on image tokens, so it cannot be applied to a CNN backbone based ViT object detectors. The BAViT model works on $384\times384$ input and YOLOS (tiny) expects $512\times512$ inputs to achieve the benchmark mAP. BAViT outputs the classification of each token as BG or FG with a total of $576$ tokens but the YOLOS model expects $1024$ tokens so we upscale the tokens labels from $576$ to $1024$ keeping the relative BG/FG patch position same. After the label scaling step, each of $1024$ token is classified as BG or FG token. The YOLOS model only computes the FG tokens from first to the final layer. We also modify YOLOS model slightly so that it does not compute anything for the BG tokens and return zeros as the final output token for these tokens. All the FG tokens are processed in the usual manner. So, the modified BAViT + YOLOS-tiny model contains 14 layers, first 2 layers of BAViT and the 12 layers of YOLOS-tiny. 

\section{Results}
\label{sec:results}
\subsection{BG/FG Classification Model}

The BAViT model was trained with both 2 layers (BAViT-small) and 10 layers (BAViT-large) depth. Table \ref{tab:models_accuracy_table} displays the token classification accuracy of these models on different datasets. BAViT-small is used for integration with object detection model (YOLOS) but we also trained the BAViT-large model to assess the impact on model accuracy. We found that BAViT-small achieved $75.93$\% accuracy, which was reasonable compared to BAViT-large’s $88.79$\% accuracy for the BG/FG classification task on VOC dataset given the difference in number of parameters for these two models. We also trained both models on COCO dataset as shown in Table \ref{tab:models_accuracy_table} and used BAViT-small trained with COCO with mAP 70.88\% as pre-processing block . YOLOS-tiny model has 6.5M parameters using 18.8 GFLOPS. Addition of BAViT-small over the native YOLOS-tiny marginally increases the total number of parameters (by 1.49M) and FLOP counts (+1.961 GFLOPS) but substantially reduced the amount of total number of tokens (25.63\%) which has the quadratic impact over the computational complexity of the ViT models. On the other hand focus DETR \cite{focus_detr} models with ResNet50 backbone has 48M parameters using GFLOPS which is almost 8x times bigger and slower. Our results also suggest that it can be applied to different datasets with configurable number of layers based on latency and RAM constraints. 

\begin{table*}[ht!]
\footnotesize
\centering
\caption{Accuracy of FG/BG classification on different models trained on different datasets and with different number of layers.}
\label{tab:models_accuracy_table}
% \begin{tabularx}{\linewidth}{p{3.5cm} | p{2cm} | p{3cm} | p{3cm}}
% \begin{tabularx}{\linewidth}{c|p{3cm}|p{3cm}|p}
% \begin{tabularx}{0.8\linewidth}{>{\raggedright\arraybackslash}p{3cm}|>{\centering\arraybackslash}X|>{\centering\arraybackslash}X|>{\centering\arraybackslash}X>}
\begin{tabularx}{0.5\linewidth}{c|c|c|c}

\toprule
\textbf{Model} & \textbf{Depth} & \textbf{Dataset} & \textbf{Accuracy(\%)} \\
\midrule

BAViT-small & 2 & Pascal-VOC  & 75.93 \\
BAViT-large & 10 & Pascal-VOC & 88.79 \\
BAViT-small & 2 & MS-COCO & 70.88 \\
BAViT-large & 10 & MS-COCO & 80.57 \\

\bottomrule
\end{tabularx}

\end{table*}

\begin{figure}
    \centering
    \includegraphics[width=1.00\linewidth]{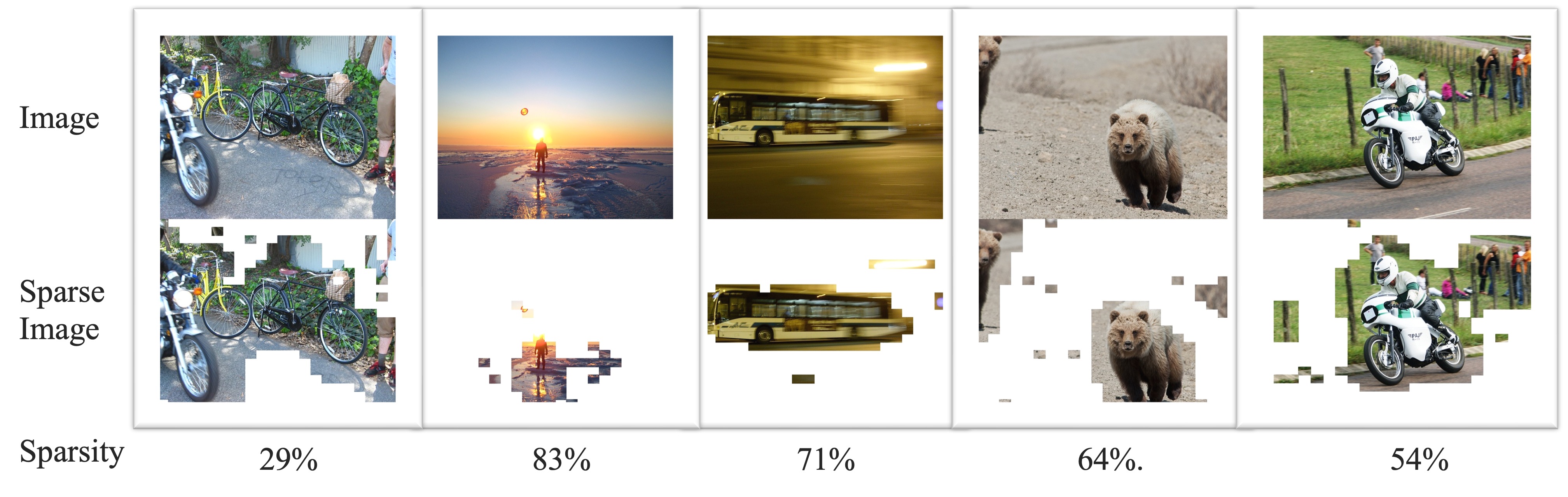}
    \caption{FG/BG token classification (16x16) on COCO images.  Top- original image, bottom - sparse image generated from BAViT with sparsity percentage.}
    \label{fig:coco_sparse_output}
\end{figure}

Figure \ref{fig:coco_sparse_output} shows the BAViT-large model output for COCO images where top image is the original image and bottom image is sparse image with all background patches shown in white color. There are few misclassified tokens where background is classified as foreground and vice versa. Foreground being classified as background is concerning so we added additional post processing block to improve the classification accuracy as explained in Appendix section.  It is evident that our model is able to separate FG/BG patches effectively even with multiple objects from different classes. It is also clear from these images that sparsity varies based on the image and it can be even more than 70\% in many images. COCO images have an average of $40$\% of background tokens which means that only $60$\% tokens are important.

\begin{table*}[ht!]
\caption{Token reduction using BAViT as a pre-processing block to YOLOS-tiny model. Total number of tokens for 2 layers of BAViT is 1152 (576 tokens per layer for $384\times384$ input ) and total number of tokens for 12 layers of BAViT is 12288 (1024 tokens per layer for $512\times512$ input). BAViT+YOLOS-F is the fine-tuned YOLOS model using only Foreground tokens (30 epochs)}
\centering
\setlength\tabcolsep{4pt}
\label{table:bavit_yolos_results}
\begin{tabularx}{0.9\linewidth}{p{3.0cm}|p{1.5cm}|p{1.5cm}|cccc|c}
\toprule
 & \textbf{} & \textbf{} & \multicolumn{5}{c}{\textbf{Number of Tokens}} \\
\midrule

\textbf{Model} & \textbf{Sparsity} & \textbf{mAP} & {} & {} & \textbf{YOLOS } & \textbf{YOLOS} & \textbf{\%Reduction}\\
{} & \textbf{\%age} & {(COCO)} & \textbf{BAViT} & \textbf{YOLOS} & \textbf{Pruned} & \textbf{+BAViT} & {}\\
\midrule

{BAViT+YOLOS} & 46\% & 20.00 & 1152 & 12288 & 6635  & 7787  & 36.63\% \\
{BAViT+YOLOS} & 43\% & 21.50 & 1152 & 12288 & 7004  & 8156  & 33.63\% \\
{BAViT+YOLOS} & 40\% & 22.50 & 1152 & 12288 & 7372   & 8524   & 30.63\% \\
{BAViT+YOLOS} & 39\% & 22.70 & 1152 & 12288 & 7495  & 8647  & 29.63\% \\
{BAViT+YOLOS} & 37\% & 23.80 & 1152 & 12288 & 7741  & 8893  & 27.63\% \\
\textbf{BAViT+YOLOS} & 35\% & \textbf{24.40} & 1152 & 12288 & 7987   & 9139   & 25.63\% \\
\textbf{{BAViT+YOLOS-F}} & 35\% & \textbf{26.60} & 1152 & 12288 & 7987   & 9139   & 25.63\% \\
{BAViT+YOLOS} & 32\% & 25.00 & 1152 & 12288 & 8355  & 9507  & 22.60\% \\
{BAViT+YOLOS} & 29\% & 25.90 & 1152 & 12288 & 8724  & 9876  & 19.60\% \\
{BAViT+YOLOS} & 5\%  & 27.70 & 1152 & 12288 & 11673  & 12825  & -4.37\% \\
{BAViT+YOLOS} & 2\%  & 28.60 & 1152 & 12288 & 12042 & 13194 & -7.38\% \\
{BAViT+YOLOS}         & 0\%  & 28.80 & 1152 & 12288 & 12288 & 13440 & -9.40\% \\

\\
\bottomrule
\end{tabularx}
\end{table*}

\subsection{Token reduction using BAViT}
As explained in section \ref{sub:bavit_integration}, we added BAViT-tiny to pre-process the image and classify each patch as FG or BG tokens before passing to YOLOS model. Using FG patches for all computation and ignoring all the BG patches, we can reduce the number of tokens in YOLOS-tiny model drastically. Equation \ref{eq:throughput} shows the calculation used to calculate the average reduction in tokens for 5000 COCO validation images. Table \ref{table:bavit_yolos_results} shows BAViT model used with different level of sparsity for token pruning and the impact on mAP due to the same. The sparsity is controlled by modifying the confidence threshold of background tokens. Our BAViT model adds extra complexity to the overall model but since this model has very low complexity and it works at much lower resolution , the overall number of token is less than the original model. for eg. the model with 34\% sparsity reduces total tokens by 24\% with an accuracy drop of 2.6\% on COCO dataset. Please note that we are not demonstrating the results of fine-tuning for most of these models. However, we have finetuned one of the model with 35\% sparsity and could improve the accuracy by 2 mAP points. It is important to note that we fine-tuned the model only for 30 epochs to improve the accuracy.

Although our method suffers a drop in mAP due to sparsification, it is still applicable to edge use cases whereas solution proposed in methods like Sparse DETR \cite{sparse_detr} and Focus DETR \cite{focus_detr} can't be used. Focus DETR, being the SOTA in token pruning field, uses ResNet50 and ResNet101 backbones to detect background tokens, which makes it impractical for edge use cases with very limited memory and computational capabilities. Also, Focus DETR proposes significant changes in the model architecture which necessitates the model to be retrained which is very expensive. BAViT on the other hand does not need model retraining, whereas to compensate the drop in mAP due to sparsification, it can be fine-tuned for lesser number epochs (30 epochs is used for our experiments). 

\begin{equation}
\label{eq:throughput}
% Equation for (Ty_i - (Tb_i + Ty_i * p)) / N
\text{Token Reduction} = \sum_{i=1}^{n} \frac{Ty_i - (Tb_i + Ty_i \cdot s)}{N}
\end{equation}

where \(Ty_i\) is total YOLOS tokens for \(i^{\text{th}}\) image, \(Tb_i\) is total BAViT tokens for \(i^{\text{th}}\) image, s is sparsity percentage in \(i^{\text{th}}\) image and N is total number of images.

\section{Conclusion}
\label{sec:conclusion}
In this work, we introduced a novel method for separating BG/FG patches in images by leveraging existing annotations from bounding boxes and segmentation maps to create localized annotations. We applied these annotations within a token classification training strategy, achieving an accuracy of up to {88.79}\% on the Pascal VOC dataset and {80.57}\% on the COCO dataset using a 10-layer transformer model. Notably, even with just 2 transformer layers, we were able to achieve over {75}\% accuracy on Pascal VOC and {70\%} on COCO dataset respectively. We also used BAViT-small model for pre-processing step to prune tokens of a YOLOS-tiny model. Our approach  could reduce the number of tokens by 25\% with a mAP drop of 3\% on COCO dataset. This drop is shown to be recovered (less than 2\% mAP drop) by sparse token finetuning by using just 30 epochs. BAViT approach is a low cost and low complexity alternative to SOTA methods like Focus DETR \cite{focus_detr} which works on large models not fitting on edge devices. Future work involves integrating our approach to YOLOS type of model to jointly train BG/FG classifier and object detector together to observe the accuracy-latency trade-off. Additionally, we also aim to achieve adaptive sparsity based on input image complexity, with a learnable threshold parameter similar to \cite{adaptive_sparse_vit}.

% ---- Bibliography ----
%%
\bibliographystyle{unsrt}
\bibliography{references}

% \newpage
\null
\thispagestyle{empty}
\newpage

\section*{Appendix}

\begin{figure}
    \centering
    \includegraphics[width=0.5\linewidth]{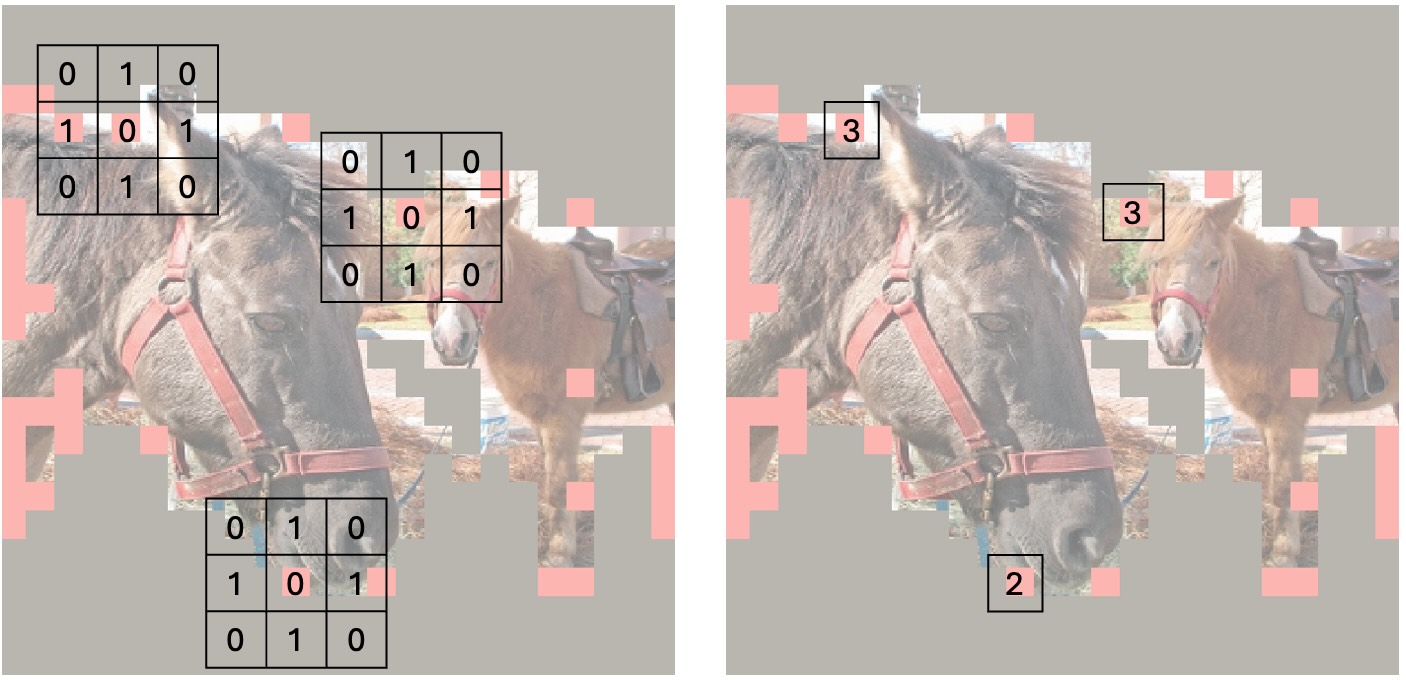}
    \caption{Converting BG tokens to FG tokens using a post processing convolution operation}
    \label{fig:merge_conv_process}
\end{figure}

\begin{figure}
    \centering
    \includegraphics[width=0.7\linewidth]{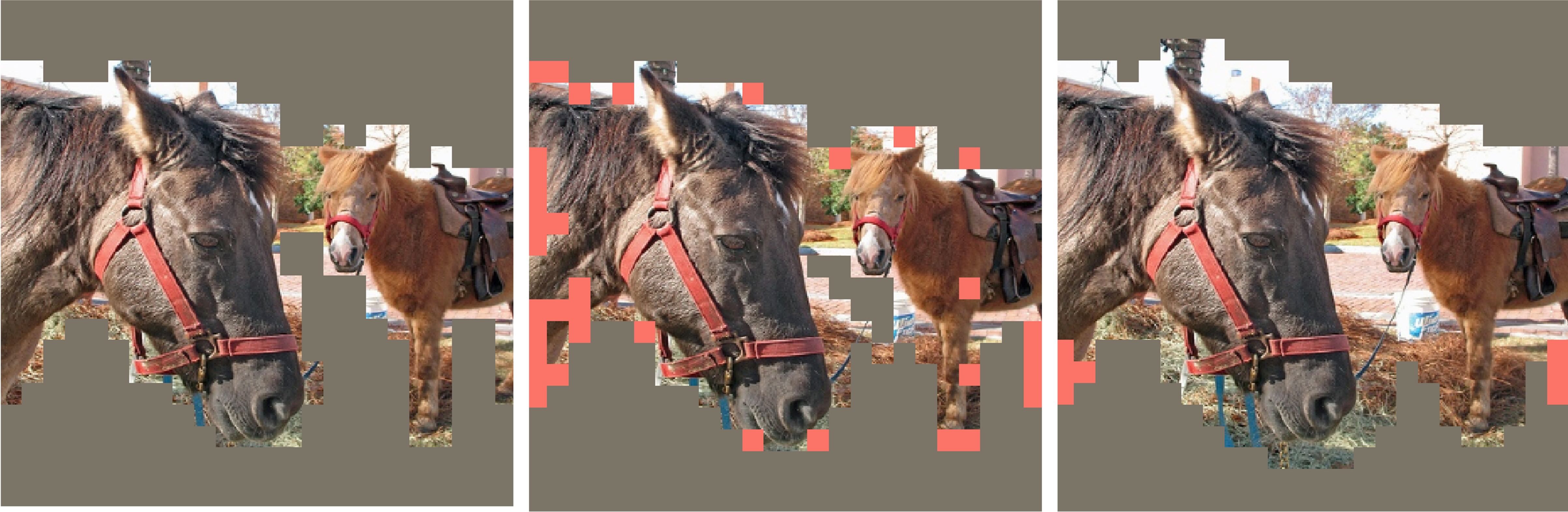}
    \caption{Left: Ground truth image (gray patch is background), Center : predicted BG/FG patches (gray patch is BG and orange patch is BG misclassified as FG, Right : Misclassified patches corrected by post-processing convolution operation}
    \label{fig:merge_conv_result}
\end{figure}

In this appendix, we provide additional details about post-processing algorithm applied to improve BG/FG classification results. Figure \ref{fig:merge_conv_result} shows the result of BAViT where orange patches are FG misclassified as BG and gray patches are correctly classified by the model. To minimize the error due to misclassified FG pixels, we use  Connected Component Analysis (CCA) \cite{he2017connected}, the traditional graph analysis algorithm to connect nodes with connected neighbors. In this case, each patch is considered as a node of the graph and CCA is performed by applying a convolutional kernel (shown in Figure \ref{fig:merge_conv_process}) on the graph (FG=1, BG=0) and converting the graph node from $0$ to $1$ for all pixels with convolution output greater than $2$. The CCA algorithm is applied for few steps to minimize the classification error. More steps reduces classification error but it increases number of FG patches which were BG in the ground truth image. We found 3 steps to be optimal based on different experiments, impact on accuracy and efficiency. Right image in Figure \ref{fig:merge_conv_result} shows the result of our post processing convolution which brings the result very close to the ground truth. Please note that, we have not applied any post processing while reporting model's accuracy in \ref{tab:models_accuracy_table} for fair evaluation. However,including the post processing convolution is expected to improve accuracy of the model significantly.

\end{document}